\documentclass{article} 
\usepackage{nips15submit_e,times}
\usepackage{hyperref}
\usepackage{url}
\usepackage{graphicx} 
\usepackage{subfigure} 
\usepackage{amsfonts}
\usepackage{caption}
\usepackage{algorithm}
\usepackage{algorithmic}

\title{Time-series modeling with undecimated\\ fully convolutional neural networks}

\author{
Roni Mittelman\\
\texttt{rmittelm@gmail.com} \\
}

\nipsfinalcopy 

\begin{document}

\maketitle

\begin{abstract}
We present a new convolutional neural network-based time-series
model. Typical convolutional neural network (CNN) architectures rely on the use of max-pooling
operators in between layers, which leads to reduced resolution at the
top layers. Instead, in this work we consider a fully convolutional
network (FCN) architecture that uses causal filtering operations, and allows for
the rate of the output signal to be the same as that of the input
signal. We furthermore propose an undecimated version of the FCN,
which we refer to as the undecimated fully convolutional neural
network (UFCNN), and
is motivated by the undecimated wavelet transform. Our experimental results verify that using the undecimated version
of the FCN is necessary in order to allow for
effective time-series modeling.  The UFCNN has several advantages
compared to other time-series models such as the recurrent neural
network (RNN) and long short-term memory (LSTM), since it
does not suffer from either the vanishing or exploding gradients problems,
and is therefore easier to train. Convolution operations can also be
implemented more efficiently compared to the recursion that is
involved in RNN-based models. We evaluate the performance of our model in a synthetic target tracking task using bearing only measurements generated from a
state-space model, a probabilistic modeling of polyphonic music
sequences problem, and a high frequency trading task using a time-series of ask/bid
quotes and their corresponding volumes. Our experimental results using
synthetic and real datasets
verify the significant advantages of the UFCNN compared to the RNN and LSTM baselines.
\end{abstract}

\section{Introduction}

Convolutional neural networks (CNNs) [1] have been successfully used to
learn features in a supervised setting, and applied to
many different tasks including object recognition [2], pose estimation [3],
and semantic segmentation [4]. Common CNN architectures introduce
max-pooling operators, which reduce the resolution at the top
layers. For tasks such as object recognition, this has the positive property of reducing the sensitivity to
small image shifts. However, in many time-series related problems such as nonlinear filtering [5], the input rate has to be
identical to the output rate, and therefore the use of the max-pooling
operators may pose a serious limitation.

In this work, we argue that allowing for the input and output layers
of CNNs to maintain the same rate, is critical for many applications
involving time-series data. Recently, fully convolutional networks
(FCNs), which allow for the input and output signals to have the same dimensions,
have been presented in the context of pixel-wise image segmentation
[11]. FCNs introduce a wavelet-transform-like deconvolution stage, which
allows for the input and output lengths to match. 

In order to apply the FCN for modeling of time-series, we propose an undecimated FCN which takes
inspiration from the undecimated wavelet transform [12], and which
replaces the max-pooling and interpolation operators with upsampling of
the corresponding filters. Our experimental results verify that this
modification is necessary in order to allow for effective time-series
modeling. We hypothesize that this is due to the translation invariant
nature of the undecimated wavelet transform. We refer to our model as the
undecimated fully convolutional neural network (UFCNN). A key feature that separates our application of
CNNs to time-series from their application to images, is that all the
convolution operations are causal. Our UFCNN model offers several
advantages compared to time-series models such as the recurrent neural
network (RNN) [10] and long short term memory (LSTM) [13],
since it does not suffer from either the vanishing or exploding
gradients problems [16,23], and is therefore easier to train. It can
also be implemented more efficiently, since it only involves
convolution operations rather than the recursion that is used in the
RNN and LSTM, and is not as straightforward to implement efficiently.

An important limitation of our model compared to the RNN and its
variants, is that it can only capture dependencies that occur within the
overall extent of the filters. Still, this can be accounted for by increasing the number of resolution levels. Since the range of the
memory that can be captured grows exponentially
with the number of resolution levels, whereas the number of parameters grows
only linearly, we argue that our model can in principle account for
both long and short-term dependencies.

We evaluate the performance of the UFCNN in three tasks. First, we
consider a toy target tracking problem of estimating the position
coordinates of the target based on bearing measurements. The second task that we consider is the probabilistic
modeling of polyphonic music using the dataset that was
presented in [22]. Finally, we evaluate the performance of the UFCNN
using a high-frequency-trading dataset of ask/bid prices
of a security, where
we use the UFCNN to learn an investment strategy. Our
experimental results show that
the UFCNN significantly outperforms the sequential importance sampling
(SIS) particle filter (for the the target tracking problem), as well as the RNN and the LSTM
baselines. 

The remainder of this paper is organized as follows. In Section 2 we
discuss related works. In section 3 we provide background on the FCNN, and
propose our undecimated FCNN. In Section 4 we present the experimental
results for the target tracking problem and for the polyphonic music dataset, and in Section 5 we present
the experimental results for learning high frequency trading
strategies. Section 6 concludes this paper.

\section{Related work}
CNNs have been introduced in [1], and have recently gained increased
popularity due to their remarkable performance in the imagenet large
scale image recognition task [2]. Aside from the availability of
larger computational resources, the other
factors which contributed to the success of CNNs are the use of the
rectified linear nonlinearity instead of a sigmoid, and the availability
of very large labeled datasets. Recent studies [14,15] have shown that
the use of very deep CNN architectures is critical in order to achieve
improved classification accuracy. 

Previous applications of CNNs to time-series signals have typically
applied the CNNs to windowed data, therefore producing a single
prediction per segment. For example, in [6,7] a CNN was used to predict the phoneme class conditional probability for a segment of raw
input signal. In [8] windowed spectrograms were used as the input to a
multi-layered convolutional restricted Boltzmann machine, and
applied to speaker identification and to audio classification tasks. CNNs
have also been used for activity recognition in videos,
however, they were mostly used to provide the feature representation
for other time-series models [9].

Another deep-learning time-series model is the recurrent neural network (RNN) [10]. RNNs can be trained using the
backpropagation-through-time algorithm [10], however, training RNNs can be a difficult task due to the vanishing and
exploding gradients problems [16,23]. There have been many
works that focussed on extending the modeling capacity of RNNs [22,26], and
alleviating the vanishing and exploding gradients problem [25]. The long
short term memory (LSTM) [13] addresses the vanishing gradients problem by
introducing additional cells that can store data indefinitely. The
network can decide when the information in these cells should be
remembered or forgotten. A multi-layered LSTM has been recently shown
to be very effective in a machine translation task [24].  

One time-series model that is particularly relevant to our proposed approach is the clockwork RNN [17], which partitions the hidden layer into different modules
that operate at different time-scales. The clockwork RNN architecture can
be interpreted as a combination of several RNNs (modules), each operating at a
different rate, where there are only inputs from RNNs that operate at lower
frequency into RNNs that operate at higher frequency. The RNNs that
operate at a lower frequencies are better suited for capturing
long-term dependencies, whereas those that operate at higher
frequency can capture the short-term dependencies. Each module is therefore
analogous to different resolution levels in our UFCNN architecture, which also operate at
different time-scales due to the appropriate upsampling of the filter coefficients.    

\section{Fully convolutional networks for time-series modeling}


\begin{figure}
\centering
  \includegraphics[scale=0.45]{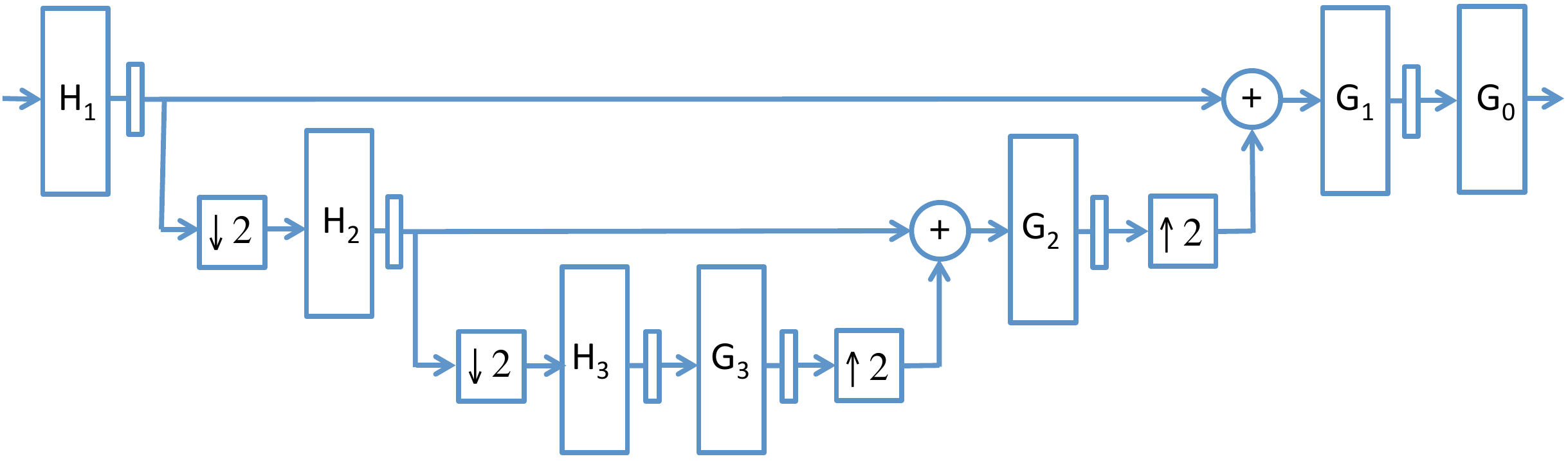}
\caption{The fully convolutional neural network with 3 resolution levels. The downsampling
  operators represent the max-pooling layers. The convolution layer
  $\mathrm{G}_0$ is used in order to allow for a negative output. The
  empty rectangles represent rectified linear units. The (+) symbols
  represent concatenation along the filters axis (``channels'' in CAFFE).}
\end{figure}

\begin{figure}
\centering
  \includegraphics[scale=0.45]{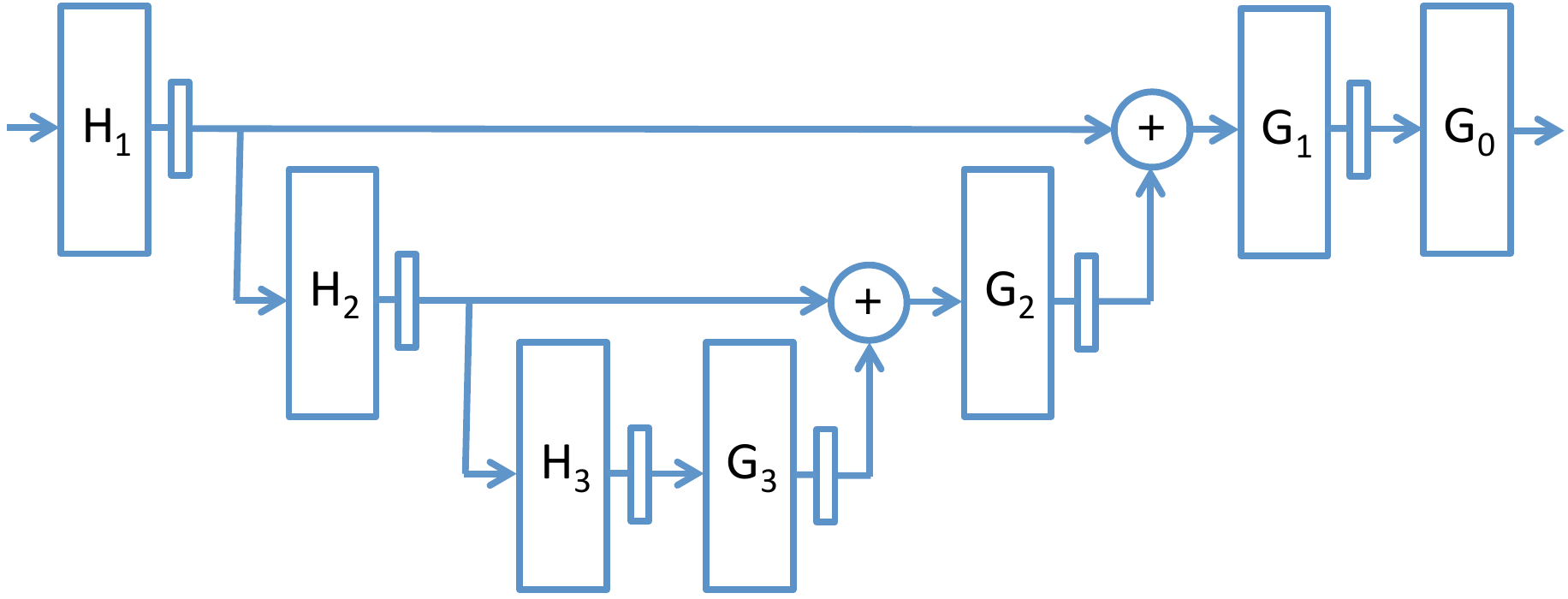}
\caption{The undecimated fully convolutional neural network with 3 resolution levels. Each
  filter $\mathrm{H}_{\ell}$ or $\mathrm{G}_{\ell}$ represents an upsampled version by a factor of $2^{\ell-1}$ of
a base filter, where $\ell$ denotes the resolution level. The (+) symbols
  represent concatenation along the filters axis (``channels'' in CAFFE).}
\end{figure}

Our proposed time-series model relies on a CNN that uses causal 1D
convolutions. In order to allow for an identical rate both for the
input and the output layers, we use a fully convolutional neural network.
In this section we first review the fully convolutional neural
network, and subsequently present our undecimated version. Our
experimental results demonstrate that using such an undecimated version is
necessary in order to allow for effective time-series modeling.

\subsection{Fully Convolutional Neural Network}
A Fully convolutional network (FCN) introduces a decoder stage that is
consisted of upsampling, convolution, and rectified linear units layers,
to the CNN architecture. The interpolated signal is then combined with
the corresponding signal in the encoder which has the same rate. This allows the network to
produce an output whose dimensions match the input dimensions. The
process is illustrated for the 1D filtering case in Figure 1, where the downsampling operations
represent the max-pooling operations. The upsampling operation doubles
the rate of the signal by introducing zeros appropriately. A FCN
was used in [11] to allow for semantic image segmentation at a finer resolution
than that which can be achieved when using standard CNNs.  

The FCN architecture is in principle very similar to a wavelet transform. As such, it also suffers from the translation variance
problem that is associated with the wavelet transform [18]. Due to the use
of the decimation operators, a small translation of the signal can
lead to very large changes to the representation coefficients. In
order to address this issue, we consider using an undecimated FCN that
is motivated by the undecimated wavelet transform [12] which is shift-invariant. 

\subsection{Undecimated fully convolutional neural networks}

The undecimated wavelet transform solves the translation variance
problem by removing the subsampling and upsampling operators, and
instead the filter at the $\ell^{th}$ resolution level is upsampled by a factor
of $2^{\ell-1}$. Our undecimated FCN therefore uses the same approach,
where the filters at the $\ell^{th}$ resolution level are upsampled by a factor
of $2^{\ell-1}$ along the time dimension, and all the other max-pooling and
upsampling operators are removed. The UFCNN is illustrated in Figure 2.


\section{Experimental results: Bearing only tracking, and
  probabilistic modeling of polyphonic music}
In this section we evaluate the UFCNN in a target tracking task, using
bearing only measurements, and using the probabilistic modeling of
polyphonic music task that
was considered in [22]. 

We begin by describing the process which we used to
generate the synthetic traget data, and then demonstrate that using
our undecimated version of the FCN instead of the ``vanilla'' FCN is
necessary in order to allow for effective time-series
modeling. Subsequently, we compare
the performance of the UFCNN to the RNN and
LSTM baselines. In all of our experiments, the filters were initialized by drawing from a zero
mean Gaussian, and are learned using stochastic gradient descent with
the RMS-prop heuristic [19] for adaptively setting the learning rate. We
implemented all the algorithms that are considered here using the CAFFE deep learning framework [20].

\subsection{Generating the target tracking data using a state-space-model}
We consider a target that moves in a bounded square, and flips the
sign of its appropriate velocity component whenever it reaches the
boundary. The observed measurement is the polar angle of the target
(with an additive noise component), where the origin
is at the center of the square. Specifically, let
$\mathrm{z}_t=[x_t;\dot{x}_t;y_t;\dot{y}_t]$, then the state
equation takes the form:
\begin{eqnarray}
\mathrm{z}_t&=&\mathrm{A}g(\mathrm{z}_{t-1})+\mathrm{w}_t,\\
\theta_t&=&\arctan{(y_t/x_t)}+\nu_t,\nonumber
\end{eqnarray}
where $\mathrm{A}=[1, 1, 0, 0;0, 1, 0, 0; 0, 0, 1, 1;0, 0, 0, 1]$,
$\mathrm{w}_t\sim\mathcal{N}(0,\textrm{diag}([0.5;0;0.5;0]\sigma_w))$,
$\nu_t\sim\mathcal(0,\sigma_{\nu})$, and
\begin{equation}
g(\mathrm{z}_t)=\left[ \begin{array}{c}
x_t\\
\dot{x}_t, \textrm{ if } -\mathrm{D}+\delta<x_t<\mathrm{D}-\delta,
                          -|\dot{x}_t|, \textrm{ if }
                          -\mathrm{D}+\delta\leq x_t, |\dot{x}_t|, \textrm{
                          otherwise } \\
y_t\\
\dot{y}_t, \textrm{ if } -\mathrm{D}+\delta<y_t<\mathrm{D}-\delta,
                          -|\dot{y}_t|, \textrm{ if }
                          -\mathrm{D}+\delta\leq y_t, |\dot{y}_t|, \textrm{
                          otherwise } \\
\end{array}\right]
\end{equation}
where $2\times\mathrm{D}$ denotes the length of the side of the square, and
$\delta$ is the radius of the target (which is assumed to be circle shaped). 

We used the hyperparameters $\mathrm{D}=10$, $\delta=0.3$, $\sigma_w=0.005$, $\sigma_{\nu}=0.005$,
for generating the data using the state-space-model. We generated
2000 training sequences, 50 validation sequences, and 50 testing
sequences, each of length 5000 time-steps. We used the standard
preprocessing approach for CNNs, where the mean that was computed using the
training set is subtracted from the input signal at each time-step.

\subsection{Evaluating the significance of the undecimated FCN}
In order to evaluate the significance of using our undecimated version
of the FCN instead of the ``vanilla'' FCN for time-series modeling, we 
used the bearings sequence to predict the target's position at each time-step using the
two versions which are illustrated in Figures 1 and 2 respectively. We
used a square loss to train the networks. In
Tables 1 and 2, we show the mean squared error (MSE) per time-step of the target's
position estimate for the two cases respectively, when varying the
number of resolution layers and the number of
filters. We used a fixed filter length of 5 for all the filters, and
trained each model for 30K iterations, starting with a base
learning-rate of $10^{-3}$ which was halved after 15K iterations, and a
RMS-prop hyper-parameter of $0.9$. We trained each model
using the training set, and we report the MSE for the validation set
(the performance on the testing set is virtually indistinguishable,
and therefore we do not report it in Tables 1 and 2).

It can be seen that using the undecimated version of the FCN significantly
outperforms the standard FCN. We hypothesize that this is due to the
shift-invariance property of the undecimated version, and the fact
that the same signal rate is maintained throughout the network.

\begin{center}
\begin{minipage}[t]{.40\textwidth}
  \centering
  \begin{tabular}{ | c | c | c | c |}
    \hline
     \# res. / \#flt. & 100 & 150 & 200 \\ \hline
    1 & 0.93 &0.86 &0.83 \\ \hline
    2 & 0.15 &0.13 &0.13 \\ \hline
    3 & 0.06 & 0.06& 0.06\\
    \hline
  \end{tabular}
  \captionof{table}{UFCNN MSE per time-step}
  \label{tab:forpol}
\end{minipage}\qquad
\begin{minipage}[t]{.40\textwidth}
  \centering
  \begin{tabular}{ | c | c | c | c |}
    \hline
    \# res. / \# flt. & 100 & 150 & 200 \\ \hline
    1 & $0.93$ &$0.86$ &$0.83$ \\ \hline
    2 & $20.14$ &$0.73$ &$4.86$ \\ \hline
    3 & $4\times 10^8$ &$1.8$ &$1.11$\\
  \hline
  \end{tabular}
  \captionof{table}{FCNN MSE per time-step}
  \label{tab:revpol}
\end{minipage}
\end{center}

\subsection{Comparison to the RNN and LSTM baselines}
The baselines to which we compare our model are the RNN and the LSTM. We used
the validation set in order to select the hyper-parameters for each
model using cross-validation. The MSE per time-step obtained for the testing set using each model is
shown in Table 3. It can be seen that the UFCNN outperforms all the
other baselines.  

\begin{table}[h!]
\centering
 \begin{tabular}{ | c | c | c | c | c |}
    \hline
           &RNN  & LSTM & UFCNN \\ \hline
    MSE  & $1.46$  & $0.09$ & $\textbf{0.06}$\\ 
  \hline
  \end{tabular}
\caption{Average MSE per time-step for the testing set of the target tracking problem.}
\label{table:3}
\end{table}

\subsection{Probabilistic
modeling of polyphonic music}
Here we evaluate the UFCNN in the task of probabilistic modeling of
polyphonic music using the ``MUSE'' and ``NOTTINGHAM'' datasets\footnote{http://www-etud.iro.umontreal.ca/~boulanni/icml2012}
[22], which include a partitioning
into training, validation, and testing sets. Each sequence
represents a time-series with a 88 dimensional binary input
vectors, representing different musical notes. We trained the models to predict the input vector at the next
time-step, using the cross-entropy loss function. The log-likelihood
obtained for the testing set using different
algorithms is shown in Table 4, where we used cross-validation to
select the hyper-parameters for each model. For the UFCNN, in both
dataset cases we used 5 resolution levels with filters
of length 2, and 50 filters at each convolution stage. It can be seen that the performance of our RNN implementation
is very close to the log-likelihood that was reported in [22] for the
same datasets. Our LSTM implementation performs (as expected) better
than the RNN, and slightly worse or similarly to the Hessian-Free optimization
RNN [25]. Our UFCNN outperforms all these baselines. We note that [22]
included comparisons to the RTRBM and RNN-RBM [22], which slightly
outperform our UFCNN. We hypothesize that this is due to the fact that
restricted Boltzmann machines (RBM) based methods tend to perform
better in the scarce data regime that applies to this case, compared
to neural network-based methods that operate better when there is
abundant data.    

\begin{table}[h!]
\centering
 \begin{tabular}{ | c | c | c | c | c | c |}
    \hline
    Dataset      & RNN (Ours) & RNN [22] & LSTM (Ours) & RNN-HF [22]& UFCNN (Ours) \\ \hline
  MUSE  & $-8.25$& $-8.13$&$-7.5$ & $-7.19$ & $\textbf{-6.67}$\\ 
  \hline
NOTTINGHAM  &$-4.62$  &$-4.46$ & $-3.85$ &$-3.89$ & $\textbf{-3.53}$ \\ 
  \hline
  \end{tabular}
\caption{Log-likelihood for the ``MUSE'' and ``NOTTINGHAM'' datasets.}
\label{table:3}
\end{table}


\section{Experimental results: Learning high frequency trading strategies}
In this section we consider the dataset of a trading competition
is available
online\footnote{http://www.circulumvite.com/home/trading-competition}. The
dataset is a time-series in which each time-step includes the best-bid-price and best-ask-price of a security,
their corresponding volumes, and several additional indicators that
might be useful for predicting future trends in the price of the
security. The rate of the time-series in the dataset is about 2-3 samples per second,
and the dataset includes a period of one year. We partitioned the data
into consecutive training, validation, and testing sets, where we used
roughly 8 months for the training set, and 2 months for each of the validation
and testing sets.

The task which we consider is to learn an investment strategy,
i.e. learn a classifier that can predict an action to be executed at
every time-step, based solely on past and present observations. The
five actions that we consider are: buy at the best-bid-price, sell at
the best-bid-price, do nothing, buy at the best-ask-price, and sell at
the best-ask-price. The learned strategy should maximize the profit
calculated using Algorithm 1,  which is based on the market
simulator that was provided in the trading competition
files. Algorithm 1 accepts as input the time-series of the best ask and bid prices and
their corresponding volumes (bidpx, askpx, bidsz, asksz), the actions taken at every time-step, as well as the maximum-position and the
cost-per-trade, which were set to 3 and 0.02 respectively in the
provided market simulator.

\subsection{Learning a trading strategy using a classification approach}
Given a time-series, we can find the optimal action to be
taken at every time-step in order to maximize the profit by combining the Viterbi algorithm and
Algorithm 1. We can then train any time-series model to predict these actions using a
softmax loss layer. A similar classification approach was used in [21] to
learn a strategy for playing Atari video games based on
offline planning. 

In Tables 5 and 6 we evaluate the UFCNN when the cost-per-trade
parameter is set to $0.02$ and $1.0$ respectively. We used cross
validation to select the hyper-parameters, where for the UFCNN we used 4 resolution
levels with filters of length 5, and the number of filters at every
convolution stage was 200 and 150 for Tables 5 and 6 respectively. We show the average
profit per time-step and the
classification accuracy for predicting the optimal decision that is
obtained using the Viterbi algorithm, for the testing set. We compare the results obtained using the UFCNN, RNN, the Viterbi
algorithm, and a uniformly random
strategy. The Viterbi algorithm is aware of the entire time-series
and serves here as the
unachievable upper bound on the performance. We trained the UFCNN
using sequences of length 5000, and we trained
the RNN model using sequences of length 200 in order to allow for faster training. It can be seen that the UFCNN is able to achieve significantly
larger returns compared to the RNN. Furthermore, it is quite close to
the unachievable upper bound that is given by the Viterbi
algorithm. For example, the UFCNN's profit in Table 5 is
50\% of the upper bound. We do not show the performance of
the LSTM since it consistently diverged during training, which
strengthens our claim that the simplified training that is offered
by our UFCNN is a major advantage compared to presently available deep-learning-based
time-series models. 

\begin{table}[h!]
\centering
\begin{tabular}{ | c | c | c | c | c |}
    \hline
           & UFCNN & RNN  & Viterbi (upper-bound) & Uniform\\ \hline
    profit/time-step & 0.13 &0.024 &0.26 &-0.01\\ 
\hline
classification accuracy &0.62  &0.38 & 1.0 & 0.2\\
  \hline
 \end{tabular}
\caption{Average profit per time-step, and classification accuracy,
  for the UFCNN, RNN, Viterbi algorithm (which serves as an upper-bound),
  and a uniformly random strategy. The cost-per-trade parameter was
  set to 0.02}
\label{table:4}
\end{table}

\begin{table}[h!]
\centering
\begin{tabular}{ | c | c | c | c | c |}
    \hline
           & UFCNN & RNN & Viterbi (upper-bound) & Uniform\\ \hline
    profit/time-step & 0.07 &0.005 &0.2 &-0.01\\ 
\hline
classification accuracy & 0.68 &0.69 & 1.0 & 0.2\\
  \hline
 \end{tabular}
\caption{Average profit per time-step, and classification accuracy, for the UFCNN, RNN, Viterbi algorithm (which serves as an upper-bound),
  and a uniformly random strategy. The cost-per-trade parameter was
  set to 1.0}
\label{table:}
\end{table}

\begin{algorithm}[t!]
\caption{Calculate profit for a length $T$ sequence}
\begin{algorithmic} 
\REQUIRE \texttt{max\_position, cost\_per\_trade, bidpx($t$),
  bidsz($t$), askpx($t$), asksz($t$), action($t$)}
\ENSURE \texttt{pnl$\leftarrow 0$, current\_position$\leftarrow 0$,
  current\_account$\leftarrow 0$}
\IF{\texttt{bidsz($1$)+asksz($1$)>0}}
\STATE \texttt{mktpx1$\leftarrow$(bidpx($1$)$\times$ asksz($1$)+askpx($1$)$\times$ bidsz($1$))/(asksz($1$)+bidsz($1$))}
\ELSE
\STATE \texttt{mktpx1$\leftarrow$(bidpx($1$)+askpx($1$))/2}
\ENDIF

\FOR{$t \leftarrow 1,\dots,T-1$}
\STATE \texttt{mktpx0}$\leftarrow$\texttt{mktpx1}
\IF{\texttt{bidsz($t+1$)+asksz($t+1$)>0}}
\STATE \texttt{mktpx1$\leftarrow$(bidpx($t+1$)$\times$ asksz($t+1$)+askpx($t+1$)$\times$ bidsz($t+1$))/(asksz($t+1$)+bidsz($t+1$))}
\ELSE
\STATE \texttt{mktpx1$\leftarrow$(bidpx($t+1$)+askpx($t+1$))/2}
\ENDIF
\STATE
\texttt{pnl$\leftarrow$pnl+current\_position$\times$(mktpx1-mktpx0)}
\IF{\texttt{action($t$)=Buy@bidpx and current\_position<max\_position}}
\STATE \texttt{current\_position$\leftarrow$current\_position+1}
\STATE \texttt{pnl$\leftarrow$pnl-(bidpx($t$)+cost\_per\_trade-mktpx1)}
\ENDIF
\IF{\texttt{action($t$)=Buy@askpx and current\_position<max\_position}}
\STATE \texttt{current\_position$\leftarrow$current\_position+1}
\STATE \texttt{pnl$\leftarrow$pnl-(askpx($t$)+cost\_per\_trade-mktpx1)}
\ENDIF

\IF{\texttt{action($t$)=Sell@bidpx and current\_position>-max\_position}}
\STATE \texttt{current\_position$\leftarrow$current\_position-1}
\STATE \texttt{pnl$\leftarrow$pnl+bidpx($t$)-cost\_per\_trade-mktpx1}
\ENDIF
\IF{\texttt{action($t$)=Sell@askpx and current\_position>-max\_position}}
\STATE \texttt{current\_position$\leftarrow$current\_position-1}
\STATE \texttt{pnl$\leftarrow$pnl+askpx($t$)-cost\_per\_trade-mktpx1}
\ENDIF
\ENDFOR
\RETURN pnl
\end{algorithmic}
\end{algorithm}

\section{Conclusions}
We presented an undecimated fully convolutional neural network (UFCNN) for
time-series modeling. Our model replaces the max-pooling and
upsampling operations that are used in the fully convolutional neural
network, with upsampling of the filters with a factor that depends on
the resolution level. Furthermore, unlike the application of
convolutional neural networks to images, we use causal filtering
operations in order to maintain the causality of the model. We demonstrated that our undecimated version of
the fully convolutional network is necessary in order to allow for
effective time-series modeling. We evaluated our UFCNN in several tasks:
bearing only target tracking, probabilistic modeling of polyphonic
music, and learning high frequency trading strategies. Our
experimental results verify that our model improves over the recurrent
neural network (RNN), and long short-term memory
(LSTM) baselines. The UFCNN has several additional advantages over
RNN-based models, since it does not suffer from the vanishing or
exploding gradients problems. It also allows for a very efficient
implementation, since it uses convolution operations that can be
implemented very efficiently.



\subsubsection*{References}


\small{

[1] LeCun Y., and Bengio, Y. (1995) Convolutional networks for images, speech,
and time-series. In M. A. Arbib, (ed.), {\it The Handbook of Brain Theory
and Neural Networks}. MIT Press.

[2] Krizhevsky, A., Sutskever, I. and Hinton, G. E. (2012)
ImageNet Classification with Deep Convolutional Neural Networks.
{\it NIPS}.

[3] Toshev, A., and Szegedy, C. (2014)
DeepPose: Human Pose Estimation via Deep Neural Networks. {\it CVPR}.

[4] Girshick, R., Donahue, J., Darrell, T., Malik J. (2014) Rich
feature hierarchies for accurate object detection and semantic
segmentation. {\it CVPR}.

[5] Arulampalam, S. M., Maskell, S., and Gordon, N., (2002) A tutorial
on particle filters for online nonlinear/non-Gaussian Bayesian
tracking. {\it IEEE Transactions on Signal Proc.}, Vol. 50, pp. 174--188.

[6] Palaz, D., Collobert, R., and  Magimai-Dos, M. (2013) Estimating
Phoneme Class Conditional Probabilities from Raw Speech Signal using
Convolutional Neural Networks. {\it INTERSPEECH}.

[7] Abdel-Hamid, O., Deng, L., Yu, D., and Jiang, H. (2013) Deep Segmental Neural Networks for Speech Recognition. {\it INTERSPEECH}.

[8] Lee, H., Largman, Y., Pham, P., and Y. Ng, A. (2009) Unsupervised
Feature Learning for Audio Classification using Convolutional Deep
Belief Networks. {\it NIPS}.

[9] Simonyan, K., Zisserman, A., (2014) Two-Stream Convolutional
Networks for Action Recognition in Videos. {\it NIPS}.

[10] Rumelhart, D. E., Hinton, G. E., and Williams, R. J. (1986) Learning representations
by back-propagating errors. {\it Nature}, 323(6088): pp. 533–536.

[11] Long, J., Shelhamer, E., and Darrell, T. (2015) Fully
Convolutional Networks for Semantic Segmentation. {\it CVPR}.

[12] Shensa, M. J. (1992) The Discrete Wavelet Transform: Wedding the
A Trous and Mallat Algorithms. {\it IEEE Transaction on Signal Processing}, Vol. 40, No. 10.

[13] Hochreiter, S., and Schmidhuber, J. (1996) Bridging long time lags by
weight guessing and ”long short term memory”. {\it Spatiotemporal
models in biological and artificial systems}.

[14] Simonyan, K., and Zisserman, A. (2015) Very Deep Convolutional
Networks for Large-Scale Image Recognition. {\it CVPR}.

[15] Szegedy, C., Liu , W., Jia, Y., Sermanet, P., Reed, S., Anguelov
, D., Erhan , D., Vanhoucke, V., and Rabinovich, A.
(2015) Going Deeper with Convolutions. {\it CVPR}.

[16] Pascanu, R., Mikolov, T., and Bengio, Y. (2013) On the difficulty of
training recurrent neural networks. {\it ICML}.

[17] Koutník, J., Greff, K., Gomez, F., and Schmidhuber, J. (2014) A
Clockwork RNN. {\it ICML}.

[18] Simoncelli, E. P., Freeman, W. T. , Adelson, E. H., and Heeger,
D. J. (1992) Shiftable Multi-Scale Transforms. {\it IEEE
  Trans. Information Theory}, Vol. 38, No. 2, pp. 587--607.

[19] Hinton, G., Srivastava, N., and Swersky, K., Neural Networks for
Machine Learning - lecture 6. Available at http://www.cs.toronto.edu/~tijmen/csc321/slides/lecture\_slides\_lec6.pdf		

[20] Jia, Y., Shelhamer, E., Donahue, J., Karayev, S., Long, J.,
Girshick, R., Guadarrama, S., and Darrell, T. (2014) Caffe:
Convolutional Architecture for Fast Feature Embedding. {\it arXiv preprint arXiv:1408.5093}.

[21] Guo, X., Singh, S., Lee, H., Lewis, R., and Wang, X. (2014) Deep
Learning for Real-Time Atari Game Play Using Offline Monte-Carlo Tree
Search Planning. {\it NIPS}.

[22] Boulanger-Lewandowski, N., Bengio, Y. and Vincent, P. (2012)
Modeling Temporal Dependencies in High-Dimensional Sequences:
Application to Polyphonic Music Generation and Transcription. {\it ICML}. 

[23] Sutskever, I., Marten, J., Dahl, G., E., and Hinton, G., E,
(2013) On the importance of initialization and momentum in deep
learning. {\it ICML}.

[24] Sutskever, I., Vinyals, O., and Le Q., (2014) Sequence to
Sequence Learning with Neural Networks. {\it NIPS}. 

[25] Martens, J., and Sutskever, I. (2011) Learning recurrent neural
networks with Hessian-Free optimization. {\it ICML}.

[26] Michalski, V., Memisevic, R., Konda, K. (2014)
Modeling deep temporal dependencies with recurrent ``grammar
cells''. {\it NIPS}.




\end{document}